\documentclass[letterpaper, 10 pt, journal, twoside]{ieeetran}

\IEEEoverridecommandlockouts                              

\usepackage{graphics} 
\usepackage{epsfig} 
\usepackage{mathptmx} 
\usepackage{times} 
\usepackage{amsmath} 
\usepackage{amssymb}  

\usepackage{mathptmx}
\usepackage{mathrsfs}
\DeclareMathAlphabet{\mathcal}{OMS}{cmsy}{m}{n}
\usepackage{booktabs}  
\usepackage{prettyref}
\usepackage{tensor}
\usepackage[caption=false, font=footnotesize]{subfig}
\usepackage{textcomp}
\usepackage{gensymb}
\usepackage[percent]{overpic}
\usepackage{array}
\usepackage{booktabs}
\usepackage{comment}

\newrefformat{fig}{Fig.~\ref{#1}}
\newrefformat{sec}{Sec.~\ref{#1}}
\newrefformat{tab}{Tab.~\ref{#1}}
\newrefformat{algo}{Algo.~\ref{#1}}
\newrefformat{eq}{Eq.~\ref{#1}}

\usepackage[usenames,dvipsnames]{color}

\title{
Target-free Extrinsic Calibration of Event-LiDAR Dyad using Edge Correspondences
}

\author{Wanli Xing$^{1,2}$, Shijie Lin$^{1,2,3}$, Lei Yang$^2$, and Jia Pan$^{1,2}$$^{\dagger}$

\thanks{$^{1}$Wanli Xing, Shijie Lin, and Jia Pan are with the Department of Computer Science, The University of Hong Kong, Hong Kong SAR, China {\tt\footnotesize wlxing@connect.hku.hk}
\endgraf
$^{2}$Wanli Xing, Shijie Lin, Lei Yang, and Jia Pan are also with the Centre for Transformative Garment Production, Hong Kong SAR, China
\endgraf
$^{3}$Shijie Lin is also with Peng Cheng Laboratory, Shenzhen, Guangdong, China \endgraf
$^\dagger$Corresponding author}

}

\begin{document}

\maketitle
\thispagestyle{empty}
\pagestyle{empty}

\begin{abstract}

Calibrating the extrinsic parameters of sensory devices is crucial for fusing multi-modal data. Recently, event cameras have emerged as a promising type of neuromorphic sensors, with many potential applications in fields such as mobile robotics and autonomous driving. When combined with LiDAR, they can provide more comprehensive information about the surrounding environment. Nonetheless, due to the distinctive representation of event cameras compared to traditional frame-based cameras, calibrating them with LiDAR presents a significant challenge. In this paper, we propose a novel method to calibrate the extrinsic parameters between a dyad of an event camera and a LiDAR without the need for a calibration board or other equipment. Our approach takes advantage of the fact that when an event camera is in motion, changes in reflectivity and geometric edges in the environment trigger numerous events, which can also be captured by LiDAR. Our proposed method leverages the edges extracted from events and point clouds and correlates them to estimate extrinsic parameters. Experimental results demonstrate that our proposed method is highly robust and effective in various scenes.

\end{abstract}

\begin{IEEEkeywords}
Calibration and Identification, Sensor Fusion, Range Sensing.
\end{IEEEkeywords}

\section{Introduction}
\IEEEPARstart{T}{he} estimation of extrinsic parameters is a crucial problem of robotic perception and forms the basis for integrating various sensory inputs such as depth, RGB images, and other data modalities. 

Light Detection and Ranging (LiDAR) is widely utilized in robotics to enable high-precision mapping and localization \cite{xu2022fast, zhang2014loam} and collision detection \cite{zhang2022generalized}. However, LiDAR data solely provides geometric information, and therefore vision sensors are frequently employed in conjunction with LiDAR sensors to address issues such as SLAM degeneracy \cite{lin2021r, lin2022r}, perception \cite{geiger2012we}, and dynamic object tracking \cite{li2021enhancing}. 

Event cameras \cite{DAVIS}, a recent novel vision sensing technology, present unique advantages over conventional frame-based cameras, including high dynamic range, resistance to motion blur, and low power consumption. This novel sensor's efficacy has been demonstrated in a diverse range of applications such as feature extraction~\cite{li2019fa}, optical flow estimation~\cite{bardow2016simultaneous}, auto focusing~\cite{lin2022autofocus}, classification \cite{9095269}, motion deblurring \cite{lin2023fast}, and visual odometry~\cite{rebecq2016evo}.

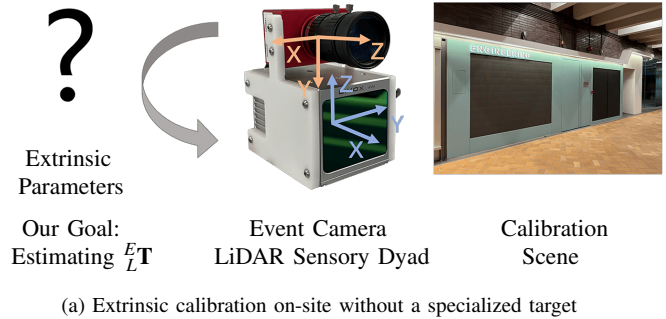
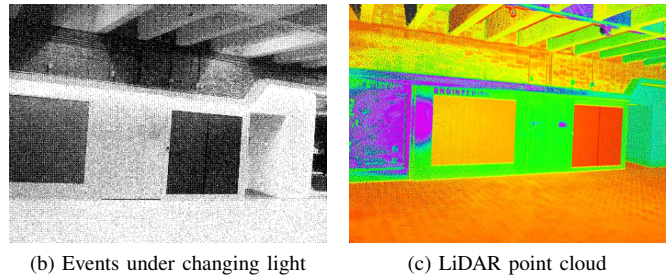
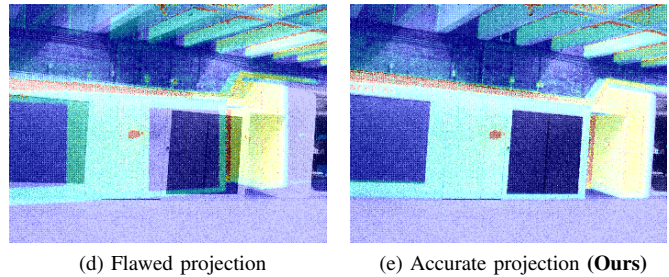
\begin{figure}[t]
    \vspace{-4mm}
    \hspace{-6.6mm}
    \subfloat[Extrinsic calibration on-site without a specialized target\label{fig:paper_purpose}]{
      \begin{overpic}[width=1.05\linewidth]{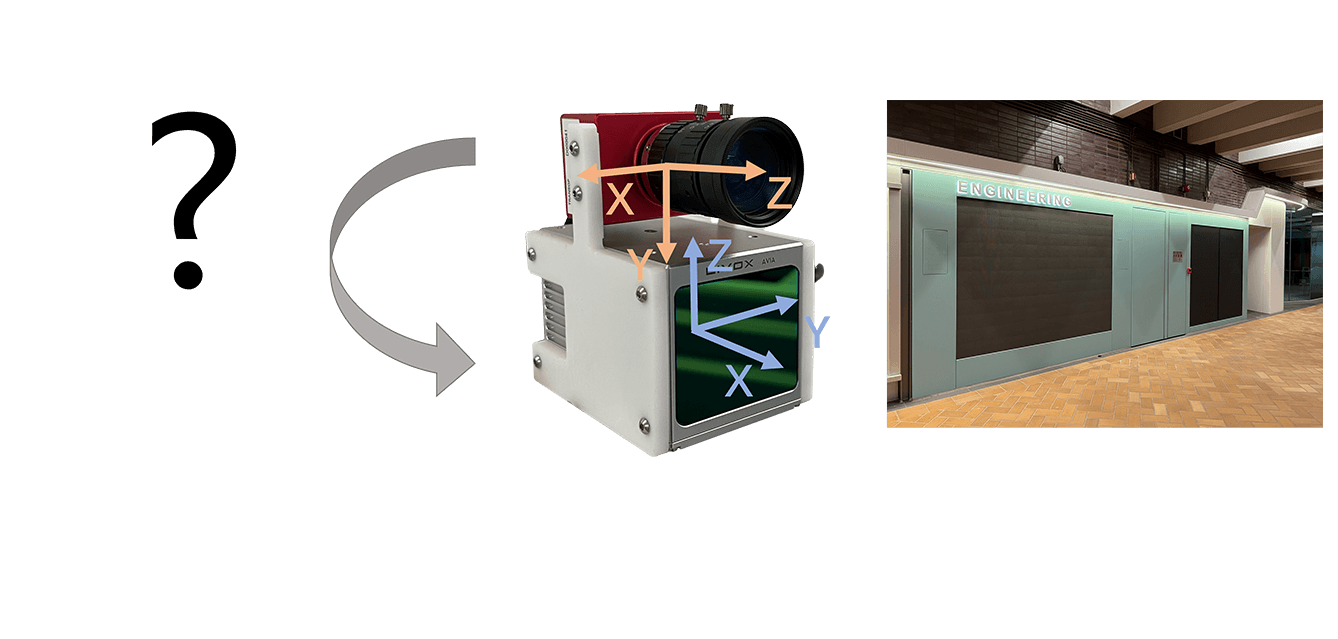}
        \put(8,17){\small Extrinsic}
        \put(7,13){\small Parameters}
        \put(7.5,7){\small Our Goal:} 
        \put(6,3){\small Estimating $\tensor*[^{E}_{L}]{\mathbf{T}}{}$}        
        \put(40,7){\small Event Camera}        
        \put(35,3){\small LiDAR Sensory Dyad}
        \put(76,7){\small Calibration}        
        \put(79,3){\small Scene}
    \end{overpic}
    \label{fig:problem}
    }
    \\
    \subfloat[Events under changing light]{%
      \begin{overpic}[width=0.475\linewidth]{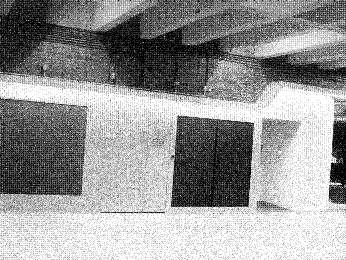}
      \end{overpic}}
    \hspace{1.8mm}
    \subfloat[LiDAR point cloud]{%
      \begin{overpic}[width=0.475\linewidth]{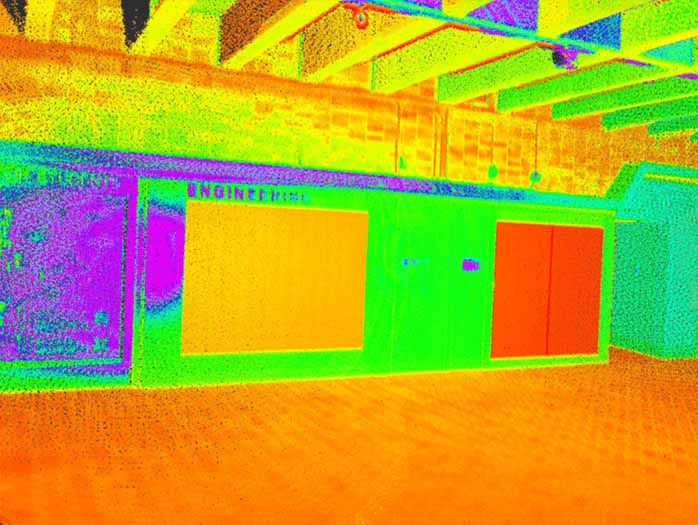}
      \end{overpic}}
    \\
    \subfloat[Flawed projection]{%
      \begin{overpic}[width=0.475\linewidth]{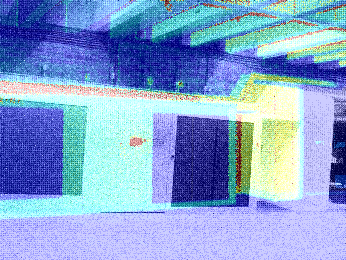}
      \end{overpic}
    \label{fig:flaw_projection}}
    \hspace{1mm}
    \subfloat[Accurate projection \textbf{(Ours)}]{%
      \begin{overpic}[width=0.475\linewidth]{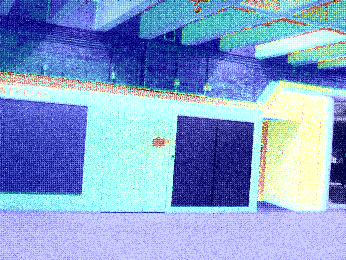}
      \end{overpic}
      \label{fig:acc_projection}
    }
    \caption{(a) In this paper, our goal is to develop a target-free method to calibrate the extrinsic transform of an event-LiDAR dyad. (b) shows an accumulated event image with a varying luminance. (c) shows a point cloud acquired by LiDAR. (d) Flawed extrinsic parameters would lead to large misalignments in projection. (e) Our estimated extrinsic parameters lead to a high-quality projection result.}
    \label{fig:bad_and_good_extrinsics}
\end{figure}

Hence, integrating LiDAR and event cameras has become an appealing approach to generate synergy for advanced sensing in robotic applications. A recent work~\cite{cui2022dense} has demonstrated promising results by fusing events and LiDAR data to generate dense depth measurements. However, integrating the two devices, i.e., the event camera and the LiDAR, presents a challenge for calibration. A custom-tailored 3D marker with a flickering display is proposed in \cite{song2018calibration} for calibrating the two devices, while an earlier work relies on the mechanical parameters for LiDAR to IMU calibration \cite{zhu2018multivehicle}. Despite the use of sophisticated devices and controlled lighting conditions, these methods struggle to produce reliable extrinsic parameter estimations for the event-LiDAR dyad, as large reprojection errors between the projected point clouds and events are observed, as illustrated in \prettyref{fig:flaw_projection}.

To promote future research on this paired system and explore its potential in robotic applications, a reliable approach for calibrating the extrinsic parameters of the paired devices must be developed. Moreover, it is desirable that the event-LiDAR dyad can be calibrated in a general environment without controlled lighting conditions or specialized equipment. 

However, two major challenges hinder the calibration process of the event-LiDAR dyad. The first challenge is the inherent difference between the data modalities. The LiDAR utilizes built-in laser emitters to sense depth, while the event camera detects brightness variations of natural light. This results in a radical difference in data format, with the LiDAR producing 3D points and the event camera generating spatiotemporal events. The second challenge is the sparsity of LiDAR data, requiring a stationary pose to obtain a dense point cloud. Meanwhile, event cameras detect brightness changes due to relative motion or lighting changes. These challenges pose significant hurdles to achieving an accurate calibration process for the event-LiDAR dyad.

\begin{figure*}[t]
\vspace{3mm}
\centering 
\includegraphics[width=0.75\linewidth]{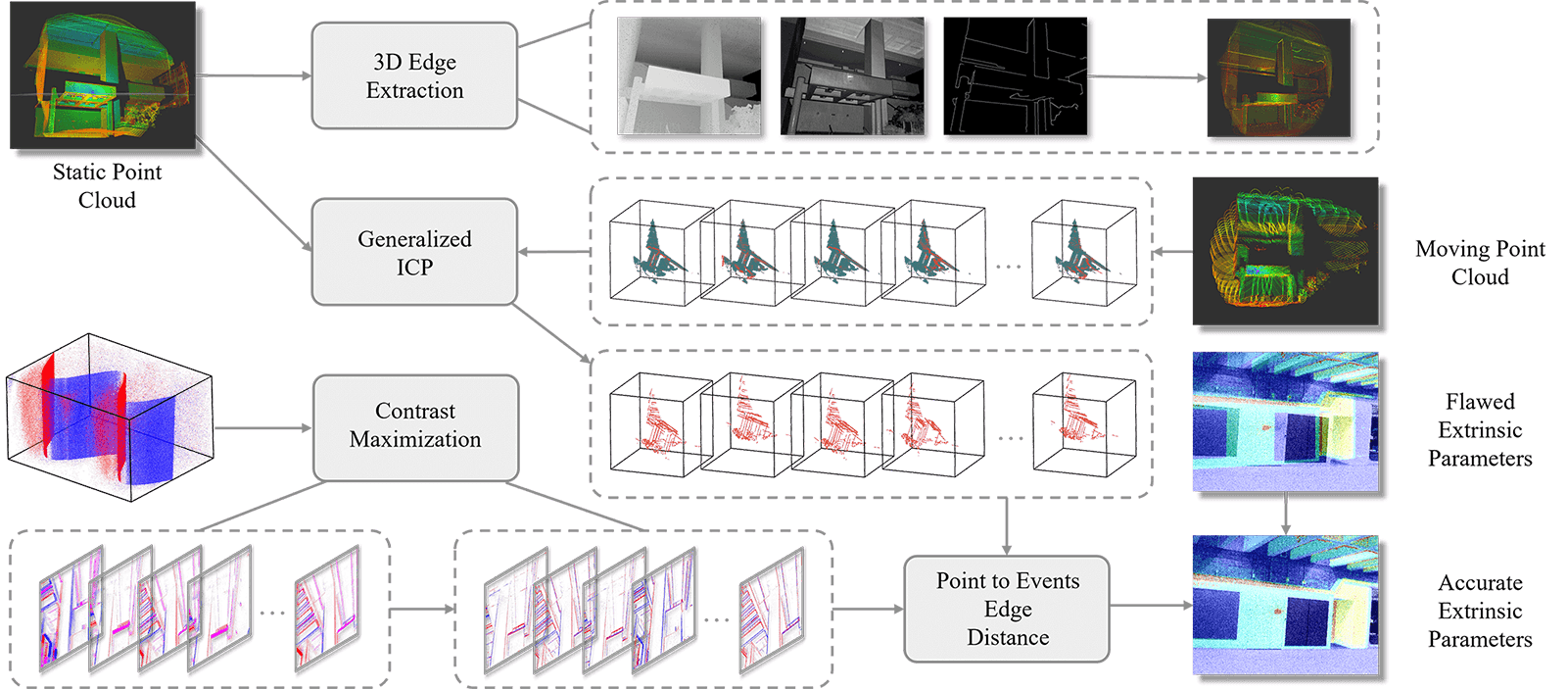} 
\caption{\textbf{Calibration pipeline.} First, we maintain the event-LiDAR dyad still to capture a dense point cloud $\mathcal{P}_s$ (top left). The edges from $\mathcal{P}_s$ are extracted using both geometry and reflectivity information from the LiDAR data. 
Then, we rotate the sensing system to capture a sequence of motion data. 
We estimate the angular velocity of the rotational motion of the sensing system and derive a set of sharp events by solving a Contrast Maximization problem based on the acquired events. 
Subsequently, we use the estimated angular velocity to undistort the (sparse) point cloud data captured in motion and register them to $\mathcal{P}_s$ via Generalized-ICP. Finally, we associate the edges extracted from the point cloud with the sharp events and solve extrinsic parameters based on the points-to-events association.
} 
\label{fig:pipeline} 

\end{figure*}

In this paper, our primary research goal is to propose a target-free approach to calibrate the extrinsic parameters of an event-LiDAR dyad using reflectivity information in LiDAR. We take inspiration from the use of reflectivity information to calibrate a LiDAR and a frame-based camera in a prior work~\cite{pandey2015automatic} and observe that the reflectivity changes in the LiDAR data match well with the patterns perceived in the event cameras (cf. Fig.~\ref{fig:edge_extraction}). This property allows us to establish a reliable correspondence between the two data streams and bridge the modality gap between events and LiDAR data to estimate the extrinsic parameters between the devices. We detect edges in the point clouds with reflectivity variations and associate them with the events captured by the event camera. The extrinsic transform is then found as the minimizer of the re-projection error between the point-cloud edges and their associated event edges.
Due to the sparsity of both sensory modalities, we develop a pipeline (\prettyref{fig:pipeline}) comprising a static stage to collect dense point cloud data followed by a moving stage to establish correspondence between the events and the point clouds across multiple poses.

In summary, our contributions are:

1) We propose an automatic extrinsic calibration pipeline for an event-LiDAR dyad, which can be used in general scenes without requiring controlled lighting or specialized targets;
 
2) We observe that geometric edges and reflectivity-changing edges present in the environment well match the events observed by the event camera, and exploit both types of edges to establish correspondence between the two data modalities;
 
3) We evaluate the consistency and accuracy of our method in multiple scenes, including indoor and outdoor scenes with complex geometry and random surface textures in various lighting conditions. We demonstrate accurate and robust extrinsic estimations in comparison with state-of-the-art methods.

\section{Related Works}

In this section, we briefly outline existing methods for calibrating both traditional frame-based cameras with LiDAR sensors and event cameras with LiDAR sensors, while discussing their primary characteristics and limitations.

\subsection{Extrinsic Calibration of Conventional Camera-LiDAR System}

Extensive research has been conducted on extrinsic calibration methods for traditional frame-based cameras and LiDAR. These methods can be classified into two categories based on whether they require a specific target. The first category comprises methods that use pre-prepared markers with known geometric parameters, such as standard planar checkerboards~\cite{zhou2018automatic,zhang2004extrinsic} or other custom targets~\cite{domhof2021joint,park2014calibration}. These methods detect the markers from camera images and LiDAR point clouds and establish the correspondence between pixels and 3D points to obtain the extrinsic parameters. The second category comprises methods that do not require specific targets. Certain methods within this category use geometric features in the environment, such as planes and edges, to solve for extrinsic parameters through mutual information or feature matching~\cite{yuan2021pixel,pandey2012automatic}. 
However, these methods are designed for traditional frame-based cameras and LiDAR and require cameras and LiDAR to remain stationary during the calibration process. Unfortunately, the imaging principle of event cameras makes these methods unsuitable for use with such cameras.

\subsection{Extrinsic Calibration of Event Camera-LiDAR System}

Theoretically, an event camera does not generate events in a stationary state with constant illumination. To address this conflict, some event camera intrinsic calibration methods~\cite{dominguez2019bio,mueggler2014event} have been designed with markers featuring blinking LEDs or screens. This allows event cameras to observe these markers in a static pose. This same concept has been applied to the extrinsic calibration of event cameras and LiDAR. Song et al.~\cite{song2018calibration} proposed a calibration marker with four circular holes and placed a blinking screen behind it. The calibration marker and the background can be distinguished in the point cloud according to the depth, and the event camera generates events corresponding to the four circular holes in the calibration marker. However, this method requires a pre-prepared calibration marker and a display, which isn't always feasible or practical. Ta~\cite{ta2022l2e} \textit{et al.} noticed that certain types of event cameras are sensitive to infrared light, enabling them to detect the brightness changes caused by LiDAR with the right wavelength. This property was then used to associate LiDAR 3D points with 2D events, and the sensors' extrinsic parameters were optimized by maximizing the mutual information between different sensory measurements. However, this method requires that the LiDAR's wavelength be within the visible spectrum of the event camera and that the LiDAR be capable of producing brightness changes that are strong enough to trigger events, which is not feasible in a general setup.

Most event-based datasets rely on indirect calibration to calibrate their LiDARs and event cameras. For instance, Gehrig \textit{et al.} \cite{gehrig2021dsec} collect the DSEC dataset using two Prophesee GEN3.1 event cameras and a Velodyne VLP-16 lidar. They first leverage the reconstructed intensity from events to calibrate the stereo event cameras and then peform point-to-plane ICP algorithm \cite{chen1992object} to align the 3D points generated by the stereo cameras with the LiDAR points to obtain extrinsic parameters. Zhu \textit{et al.} \cite{zhu2018multivehicle} collect the MVSEC dataset using two DAVIS346 event cameras and a Velodyne VLP-16 LiDAR. They use the standard image output of DAVIS 346 and conduct standard calibration with the Camera and Range Calibration Toolbox \cite{geiger2012automatic} and then fine-tune the result manually.

\begin{figure}[t]
\centering
\begin{overpic}[width=\linewidth]{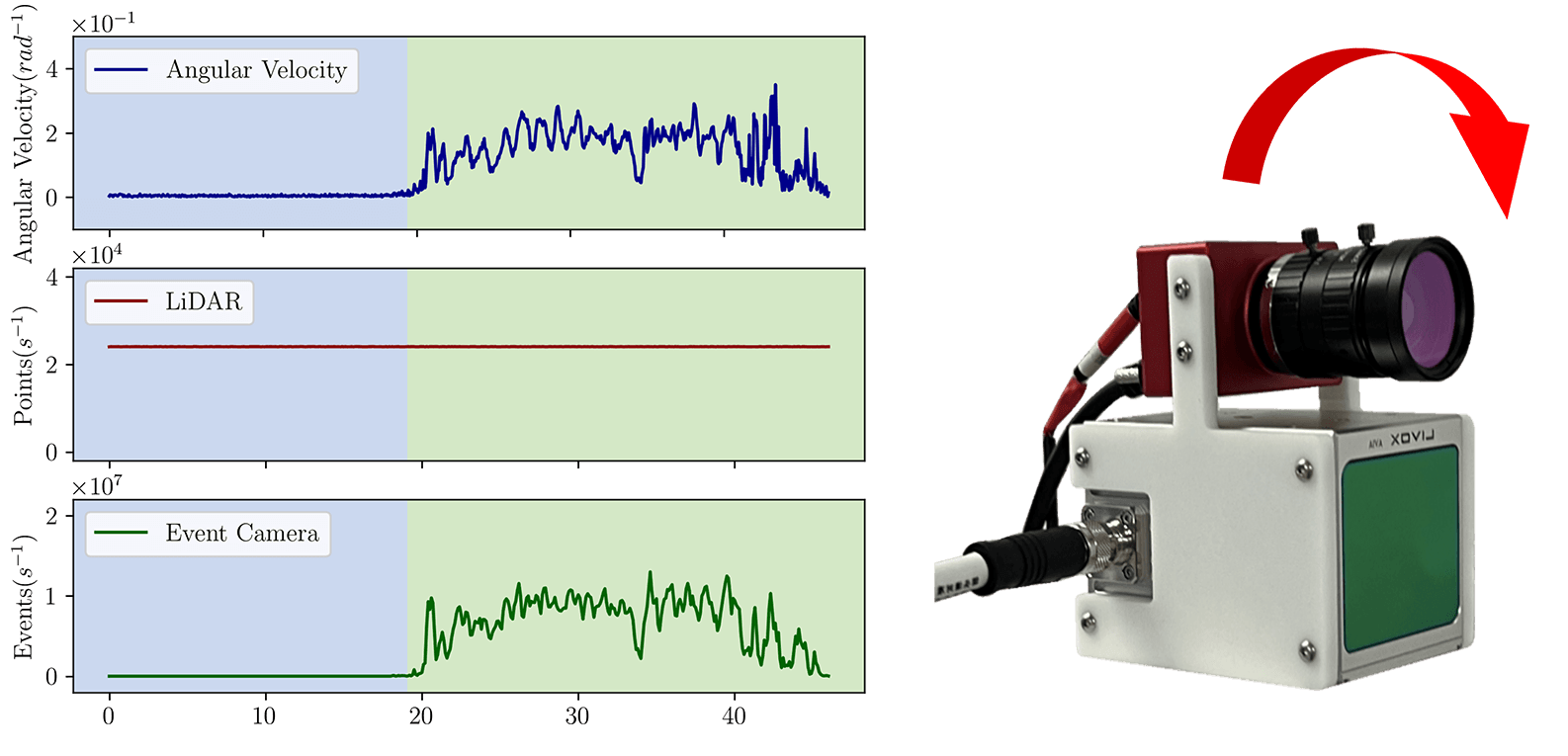}
\put(6,-1){\small $t_0$}
\put(25.5,-1){\small $t_1$}
\put(54,-1){\small $t_n$}
\end{overpic}
\caption{The event-LiDAR dyad keeps static from $t_0$ to $t_1$ to accumulate a dense point cloud and then move from $t_1$ to $t_n$ to collect events. Due to the imaging property of event cameras, the event rate is low when the sensor is stationary.}
\label{fig:motion_and_events_rate}

\end{figure}

\section{Methodology}

\subsection{Overview}

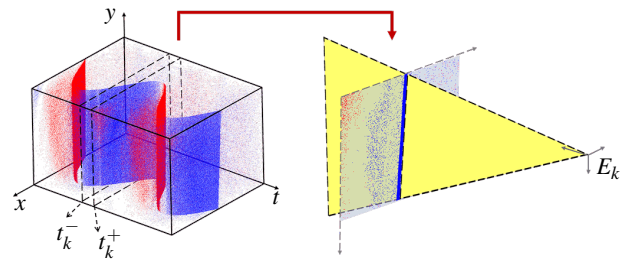
\begin{figure}[t]
\centering
\begin{overpic}[width=0.9\linewidth]{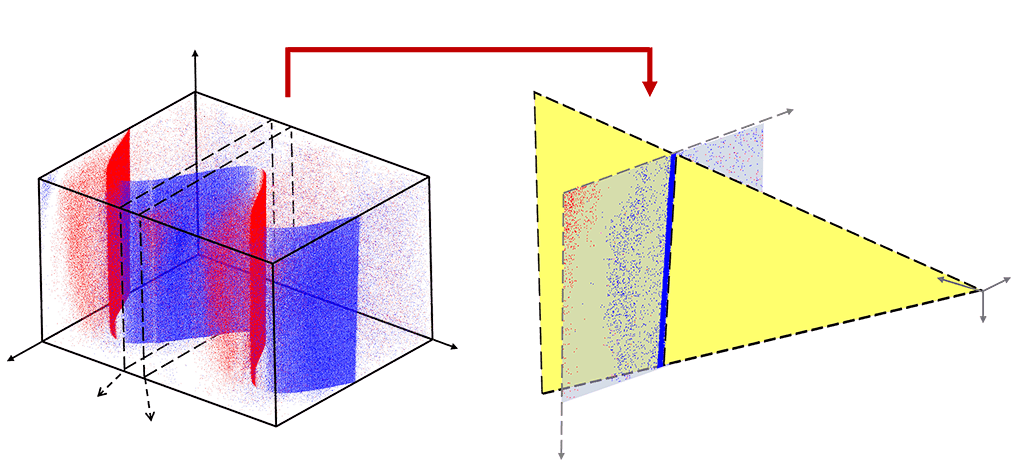}
\put(1,8){\small $x$}
\put(16,40){\small $y$}
\put(44,9){\small $t$}
\put(15,1.5){\small $t_k^+$}
\put(8,3){\small $t_k^-$}
\put(98,14){\small $E_k$}
\end{overpic}

\caption{Upon movement of the event camera, a significant amount of events are triggered by the 3D edge in space. The resulting spatial-temporal distribution of these events is displayed on the left. Notably, the projection relationship between the triggered events and the 3D edge is upheld, as demonstrated on the right.}
\label{fig:constraint}

\end{figure}

In this paper, our goal is to determine the extrinsic transform, denoted as $^{L}_{E}\mathbf{T}$, between an event camera and a LiDAR in a target-free manner, without the need for specialized equipment, calibration boards, or controlled lighting conditions. To achieve this goal, we propose a pipeline, as depicted in \prettyref{fig:pipeline}, consisting of two stages: a stationary stage (from $t_0$ to $t_1$ in \prettyref{fig:motion_and_events_rate}) and a motion stage (from $t_1$ to $t_n$ in \prettyref{fig:motion_and_events_rate}). During the stationary stage, the Lidar-Event Camera pair remains stationary in the environment, capturing a dense point cloud. In contrast, during the motion stage, the pair is rotated to trigger a significant number of events that correspond to the edges of the scene, as shown in \prettyref{fig:constraint}. By formulating an optimization problem that matches the point cloud edges to the events, we are able to recover the extrinsic parameters.

Based on the observation that both geometric and
reflectivity-changing edges present in the environment
can well match the events triggered when the event camera moves, we first describe how to detect these edges from the static point cloud acquired by the LiDAR in~\prettyref{sec:subsec_pointcloud_edges_extraction}.  

In~\prettyref{sec:contrast_maximization} we introduce a method based on Contrast Maximization to estimate the angular velocity of the event-LiDAR dyad when it moves. With this, we can warp the accumulated event images to obtain sharp event edges and undistort the (sparse) point cloud acquired during motion. The undistorted point clouds can then be registered with the static point cloud to obtain dense geometric and reflectivity-changing edges.

In~\prettyref{sec:optimize}, we present how to associate the point-cloud edges to the sharp event edges at a particular frame, as well as the formulation based on the reprojection error that solves the extrinsic transform $^{L}_{E}\mathbf{T}$.

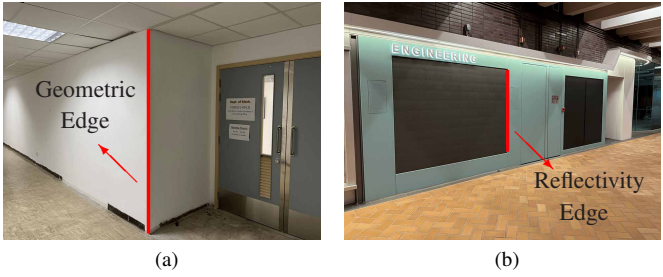
\begin{figure}[t]
    \subfloat[\label{fig:two_edges_a}]{%
      \begin{overpic}[width=0.475\linewidth]{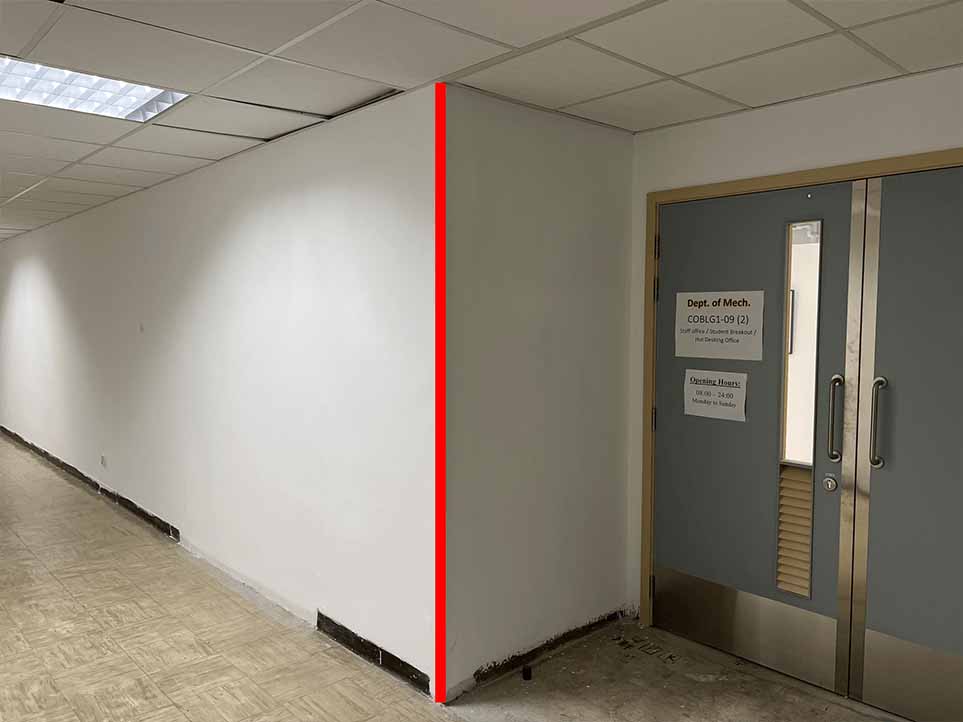}
        \put(10,45){\textcolor{Black}{\small Geometric}}
        \put(18,35){\textcolor{Black}{\small Edge} }
        \put(42,18){\textcolor{red}{\vector(-1,1){12}}}
      \end{overpic}
    }\hfill
    \subfloat[\label{fig:two_edges_b}]{%
      \begin{overpic}[width=0.475\linewidth]{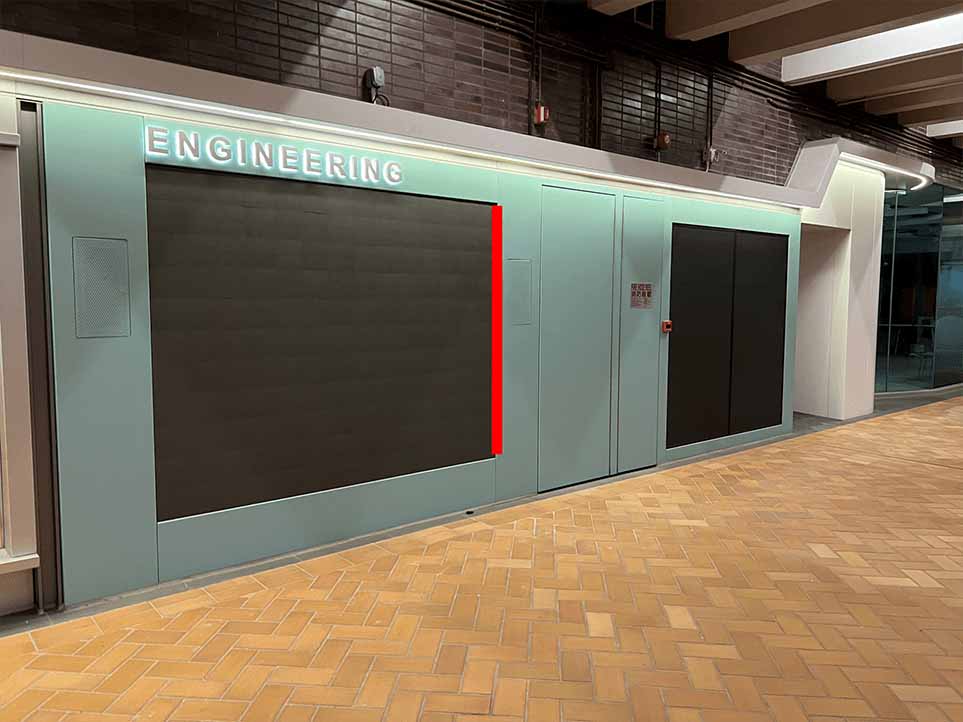}
        \put(60,16){\textcolor{Black}{\small Reflectivity}}
        \put(68,6){\textcolor{Black}{\small Edge} }
        \put(54,34){\textcolor{red}{\vector(1,-1){12}}}
      \end{overpic}
    }
    \caption{Typical examples of two types of edges. (a) A geometric edge due to different reflecting angle; (b) A reflectivity edge due to material difference.}\label{fig:two_types_of_edges}
\end{figure}

\subsection{Edges Extraction from Static Point Cloud}
\label{sec:subsec_pointcloud_edges_extraction}

\begin{figure*}[t]
    \centering
    \vspace{3mm}
  \subfloat[Depth image]{%
       \includegraphics[width=0.16\linewidth]{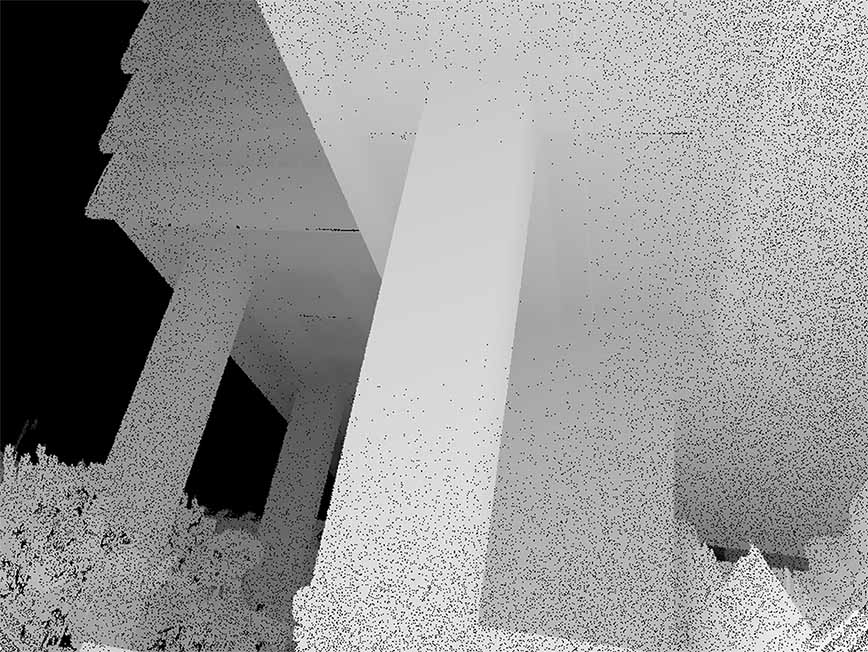}
        \label{fig:depth_img} }
        \hspace{-1mm}
  \subfloat[Reflectivity image]{%
        \includegraphics[width=0.16\linewidth]{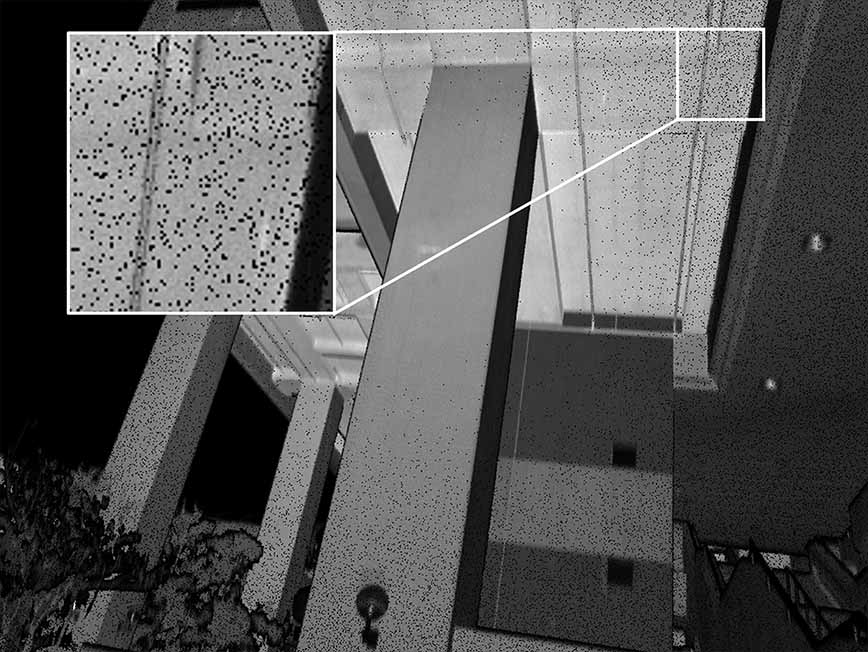}
        \label{fig:intensity_img} }
        \hspace{-1mm}
  \subfloat[Extracted 2D edges]{%
        \includegraphics[width=0.16\linewidth]{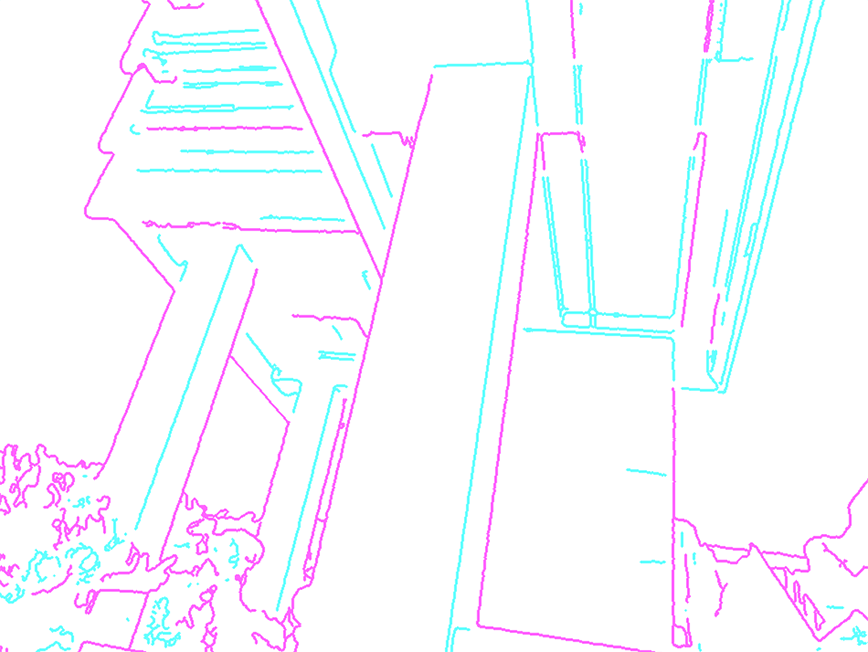}
        \label{fig:detection2d}}
        \hspace{-1mm}
  \subfloat[Extracted 3D edges]{%
        \includegraphics[width=0.16\linewidth]{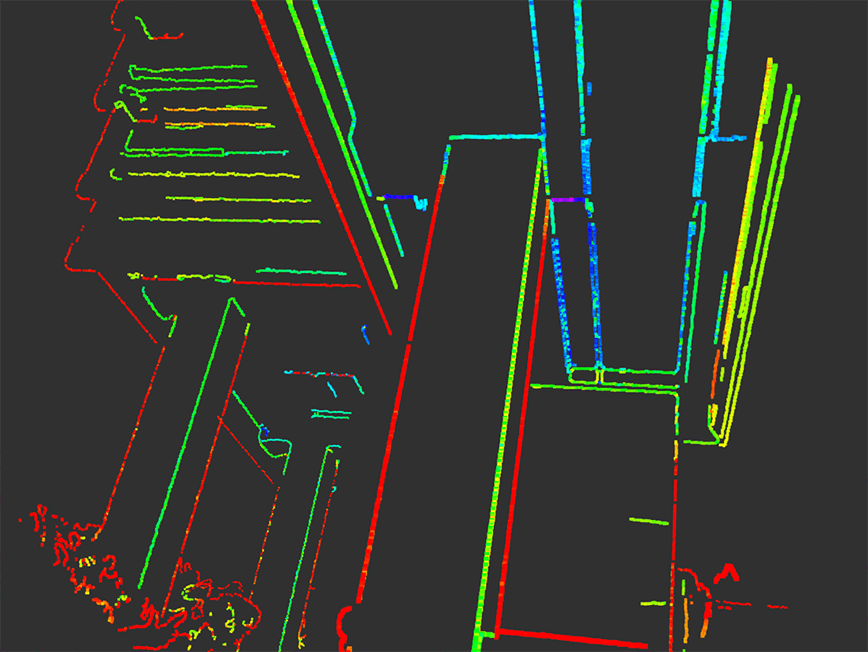}}
        \hspace{0.1mm}
  \subfloat[Sharp event image]{%
        \includegraphics[width=0.16\linewidth]{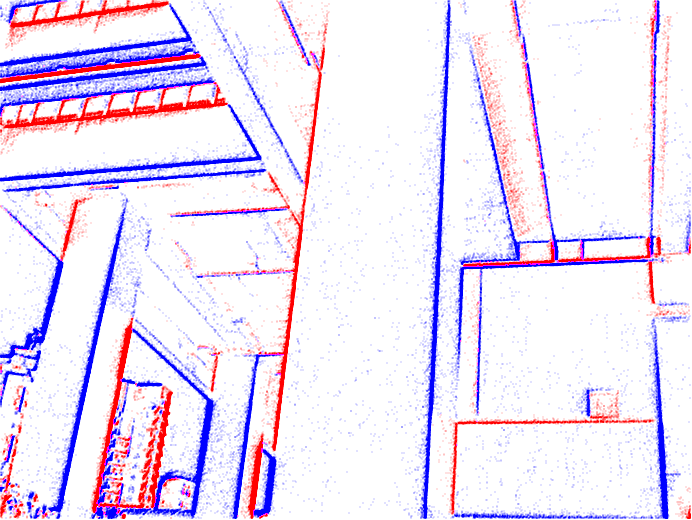}
        \label{fig:sharp_events_image}}
        \hspace{-1.6mm}
  \subfloat[Blurred event image]{%
        \includegraphics[width=0.16\linewidth]{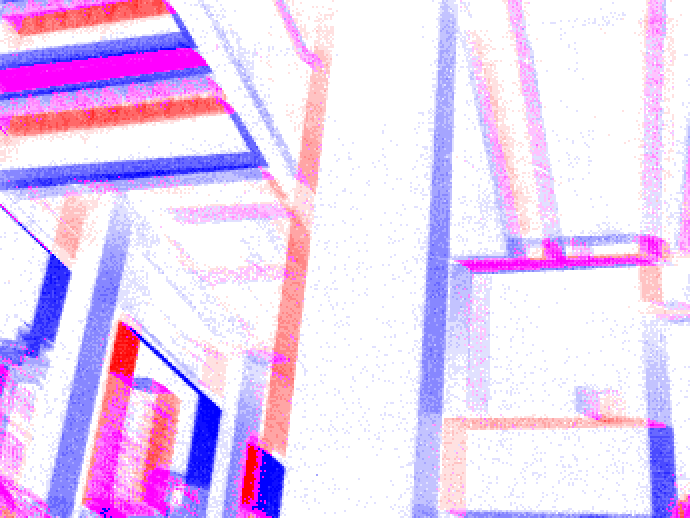}
        \label{fig:blurred_events_image}}
  \caption{The 2D edges are first extracted from (a) and (b), respectively, and then filtered to obtain edges in (c), where purple and blue represent geometric and reflectivity edges, respectively. (d) The corresponding 3D points that form the edges in the point cloud are identified through projection. With the Contrast Maximization framework, we can estimate motion parameters $\mathbf{\omega}_k$ and warp events to a sharp event image (e) from the blurred event image (f) directly accumulated during movement. }
  \label{fig:edge_extraction} 

\end{figure*}

 
We observe that event-triggering edges can be classified into two categories, namely the geometric edges and the reflectivity edges, as labeled in \prettyref{fig:two_types_of_edges} and shown in~\prettyref{fig:edge_extraction}
As the brightness on either side of these edges differs due to the angle between them, geometric edges can trigger a large number of events as the event camera moves. The reflectivity edges similarly arouse many events because the reflectance values of the materials on both sides of the edges are different. 
Since geometric edges alone are too sparse for establishing correspondences with the dense events, we aim to use the reflectivity edges as well, which are also the major trigger of events.

\begin{figure}[t]
    \subfloat[\label{fig:lidar_edge_reflectivity_a}]{%
      \begin{overpic}[width=0.475\linewidth]{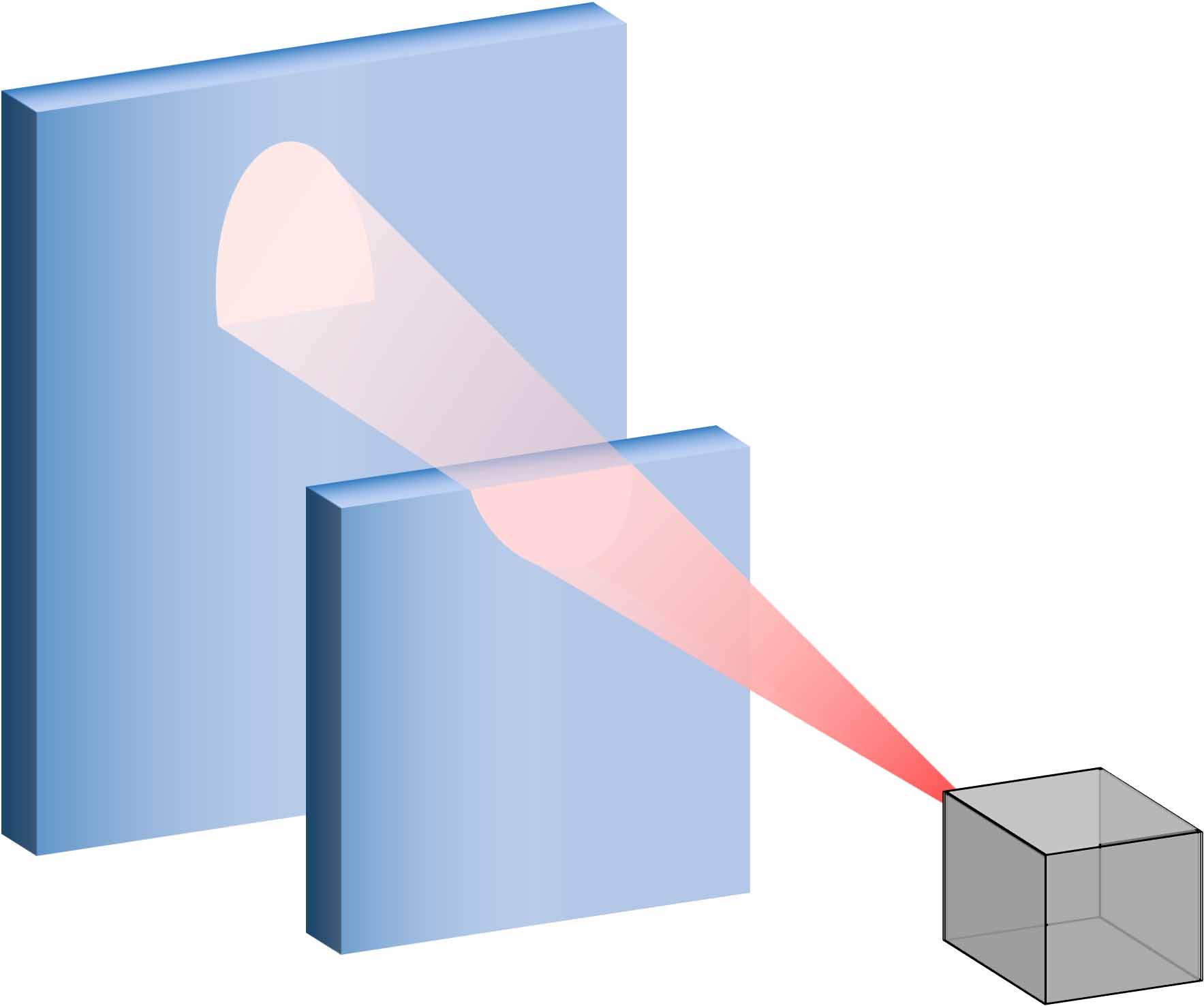}
        \put(62,56){\small Divergence}
        \put(72,48){\small Angle}
        \put(55,36){\vector(1,1){10}}

      \end{overpic}
    }\hfill
    \subfloat[\label{fig:lidar_edge_reflectivity_b}]{%
      \begin{overpic}[width=0.475\linewidth]{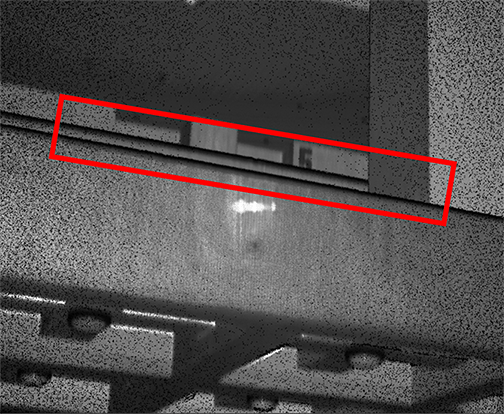}
      \end{overpic}
    }
    \caption{Incorrect reflectivity due to occlusion. (a) The divergence angle of LiDAR laser; (b) Low reflectivity area.}
    \label{fig:lidar_edge_reflectivity}
\end{figure}

In order to extract both types of edges from a point cloud, it is necessary to use the reflectivity information of the point cloud obtained by LiDAR.
However, the reflectivity information of the LiDAR data can be unreliable when the laser beam emitted from the LiDAR hit two objects with one partially occluded by the other; see \prettyref{fig:lidar_edge_reflectivity_a} as an example.
This is because the laser beam has a divergence angle, and when an object in the environment is occluded by another, the laser beam is partially reflected by both of the objects. 
In such cases, the reflectivity and depth of the point are determined by calculating the energy and time of flight of the first echo, respectively. However, since the first echo accounts for only part of the echo energy, the calculated reflectivity will be smaller than the case if the reflectivity is computed based on the full energy in the general case. 
This decrease in reflectivity can result in rapid reflectance changes around the objects as depicted in \prettyref{fig:lidar_edge_reflectivity_b} and lead to spurious edges detected by the Canny edge detector.
Note that, however, the depth estimation is not influenced by the occlusion present in the environment and thus is reliable.

Drawing from this observation, we first project the 3D point cloud $\mathcal{P}_s$, which was acquired during the stationary stage, onto a virtual imaging plane of the LiDAR as illustrated in \prettyref{fig:depth_img} and \prettyref{fig:intensity_img}. 
This is followed by employing a median filter to fill in any empty pixels due to the sparsity nature of the LiDAR device we use (i.e., Livox Avia). 
Next, we apply the Canny edge detector~\cite{canny1986computational} to the reflectivity image to detect reflectivity edges. Owing to the aforementioned problem, spurious edges could exit. Since the depth estimation is reliable in these occluded cases, we make use of the depth information to filter these spurious edges in the reflectivity image. 
To this end, we also apply the Canny edge detector to the depth map and prioritize the depth edges over those detected on the reflectivity image. Specifically, if there are edges detected on the depth map in the surrounding area ($5\times5$ pixel patch) of an edge detected on the reflectivity image, we neglect the latter from the final detection result as shown in \prettyref{fig:detection2d}.
Finally, we lift the 2D edges detected to 3D using the inverse projection to find the corresponding 3D points in the point cloud. 
When a 2D edge point corresponds to multiple 3D edge points, we calculate the center of these 3D points and consider it as the 3D edge point corresponding to this 2D edge point. We denote by $\mathcal{P}_e$ the set of 3D edge points found from the point cloud $\mathcal{P}_s$.

\subsection{Recovery of Distinct Events under Motion}
\label{sec:contrast_maximization}


The event camera outputs an event in an asynchronous manner whenever the logarithmic value of a pixel's brightness change exceeds a specified constant. A single event is described by a tuple, $e_i=(\mathbf{x}_i,t_i,p_i)$, where $\mathbf{x}_i$ is the 2D coordinate of the pixel, $t_i$ is the timestamp when an event is triggered, and $p_i$ is a binary value that represents the sign of the brightness change, colored with red and blue in \prettyref{fig:blurred_events_image}.



We use the notation $[t_k^-, t_k^+]$ to refer to a time interval centered at $t_k$ with a short duration of $\Delta t$. The events that occur within this time interval are denoted by the set $\mathcal{E}_k=\{e_i\}_{i=0}^{N-1}$. We accumulate the events in $\mathcal{E}_k$ to obtain the event image $I_k(\mathbf{x})$ in \prettyref{fig:blurred_events_image}, which is given by 
\begin{equation}
    I_k(\mathbf{x})=\sum_{i=0}^{N-1}p_i\delta(\mathbf{x}-\mathbf{x}_i),
    \label{eq:blurred_event_image}
\end{equation}
where $\delta$ is the Dirac delta function.

Event image $I_k(\mathbf{x})$ is usually blurred due to the continuous motion of the event camera in time interval $[t_k^-, t_k^+]$.
During this period, the motion of the camera can be approximated by a pure 3D rotation and the angular velocity $\mathbf{\omega}_k\in\mathbb{R}^3$ can be estimated. 
We employ the Contrast Maximization framework~\cite{gallego2017accurate,gallego2018unifying} to recover distinct events triggered by edges from the obtained event image. An event $e_i$ triggered at time $t_i$ can be warped to time $t_k$, given by
\begin{equation}
    \mathbf{x}_i^{\prime}=\exp(\hat{\mathbf{\omega}}_k(t_i-t_k))\mathbf{x}_i,
    \label{eq:rotation_warp}
\end{equation}
where $\hat{\omega}_k$ represents the cross-product matrix of ${\omega}_k$. By applying this transformation, we obtain the warped event image:
\begin{equation}
    I(\mathbf{x},\mathbf{\omega}_k)=\sum_{i=0}^{N-1}p_i\delta(\mathbf{x}-\mathbf{x^{\prime}}_i).
    \label{eq:rotation_warp_image}
\end{equation}


To estimate the angular velocity $\mathbf{\omega}_k$, an objective function $f(\mathbf{\omega}_k)$ is built from the warped events and defined as the variance of the warped event image:
\begin{equation}
\newcommand{\Var}{\mathrm{Var}}
    \begin{aligned}
       f(\mathbf{\omega}_k)
       =\Var(I(\mathbf{x},\mathbf{\omega}_k))
       =\frac{1}{|\Omega|}\int_{\Omega}(I(\mathbf{x},\mathbf{\omega}_k)-\mu_I)^2 d\mathbf{x},
    \end{aligned}
    \label{eq:cm_cost_func}
    \vspace{-3mm}
\end{equation}
where $\mu_I= \frac{1}{|\Omega|}\int_{\Omega}I(\mathbf{x},\mathbf{\omega}_k)d\mathbf{x}$ is the mean of the warped event image, computed as the average intensity over the image plane domain $\Omega$. The functional measures the contrast of the event image with a different angular velocity $\mathbf{\omega}_k$. 
The maximizer $\mathbf{\omega}_k^*$ of this functional can lead to a warped event image $I_k=I(\mathbf{x},\mathbf{\omega}_k^*)$ that contains distinct edges, as shown in \prettyref{fig:sharp_events_image}.

\subsection{LiDAR Self-motion Undistortion and Localization}
\label{sec:icp}

We aim to obtain the LiDAR's pose at time $t_k$ so as to exploit the extracted 3D edge points $\mathcal{P}_e$ to establish the correspondence with the sharpened event image at $t_k$. 
However, the acquired point cloud $\mathcal{P}_k$ at $t_k$ is subject to motion distortion as the LiDAR moves during this time interval $[t_k^-, t_k^+]$.
The LiDAR's motion, denoted as $\tensor*[^{L}]{\mathbf{T}}{^{t_k^+}_{t_k^-}}$, can be represented by the event camera's motion $\tensor*[^{E}]{\mathbf{T}}{^{t_k^+}_{t_k^-}}$ in the same time interval and the extrinsic transform $^{L}_{E}\mathbf{T}$,

\begin{equation}
    \begin{aligned}
    \tensor*[^{L}]{\mathbf{T}}{^{t_k^+}_{t_k^-}}
    =\tensor*[^{L}_{E}]{\mathbf{T}}{} \cdot
    \tensor*[^{E}]{\mathbf{T}}{^{t_k^+}_{t_k^-}}
    =\tensor*[^{L}_{E}]{\mathbf{T}}{}
        \begin{bmatrix}
        \exp(\hat{\mathbf{\omega}}_k\Delta t) & \mathbf{0}\\\
        \mathbf{0} & 1
        \end{bmatrix}.
    \end{aligned}
    \label{eq:motion_compensation1}
\end{equation}

To compensate for the motion of a point $P_j\in \mathcal{P}_k$ with the timestamp $t_j \in [t_k^-, t_k^+]$, we compute the transformation at $t_j$ with $\tensor*[^{L}]{\mathbf{T}}{^{t_k^+}_{t_k^-}}$ and the time difference $t_j - t_k$ by linear interpolation.
This way, the point cloud collected during the time interval $[t_k^-, t_k^+]$ can be undistorted and accumulated at $t_k$. We denote these undistorted point clouds as $\mathcal{P}_k^{\prime}$. Finally, we solve the transformation between $\mathcal{P}_k^{\prime}$ and $\mathcal{P}_s$ by the Generalized-ICP algorithm~\cite{segal2009generalized} and the obtained transformation at $t_k$ is denoted as $\tensor*[^{L}]{\mathbf{T}}{^{t_k}_{t_0}}$.


\begin{figure}[t]
\centering
\begin{overpic}[width=0.35\linewidth]{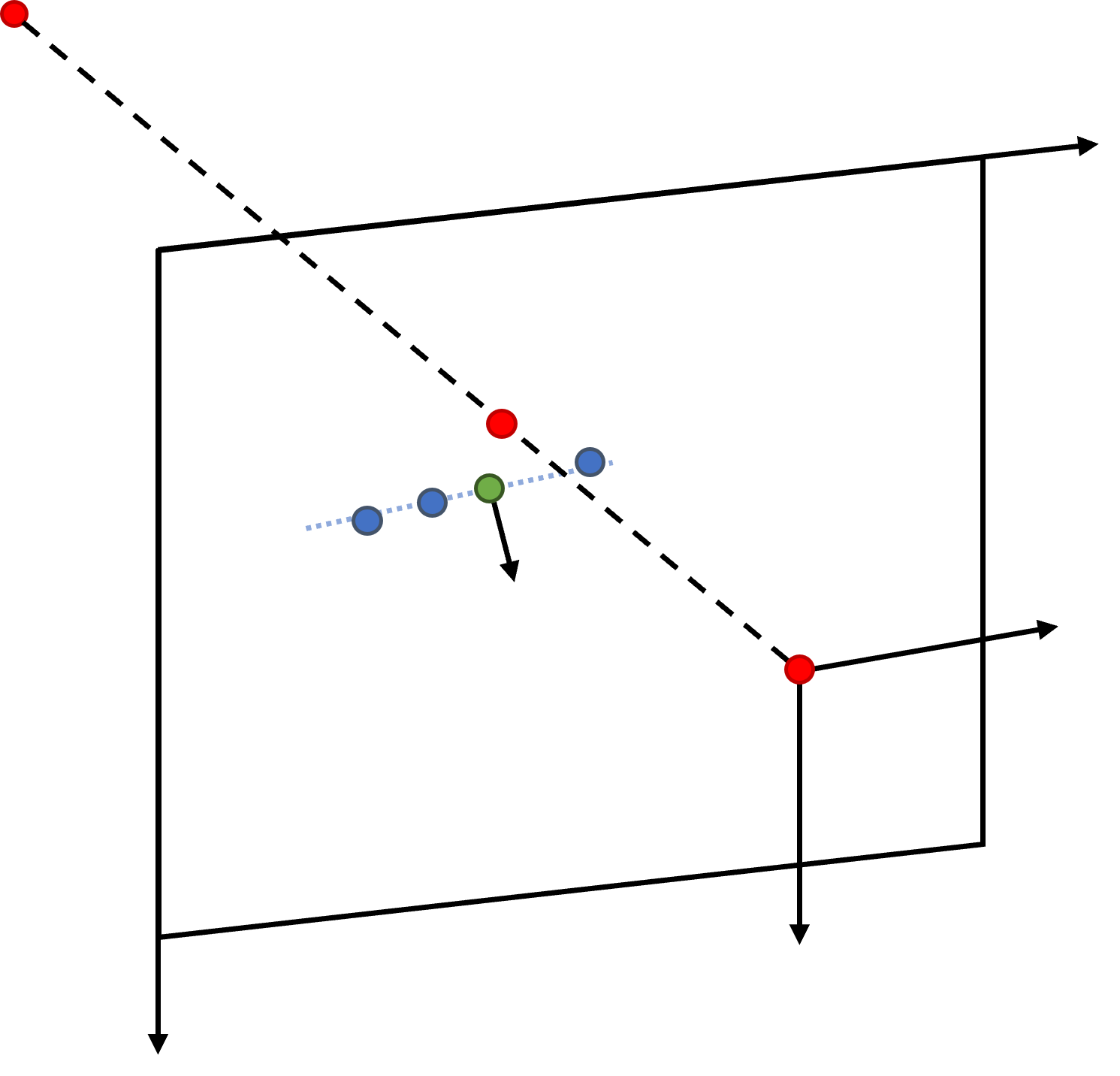}
\put(1,95){\small $\tensor*[^{L}]{\mathbf{P}}{_{i}}$}
\put(44,60){\small 
    $
    \tensor*[^{E}]{\mathbf{K}}{} \cdot
    \tensor*[^{L}_{E}]{\mathbf{T}}{^{-1}} \cdot
    \tensor*[^{L}]{\mathbf{T}}{^{t_k}_{t_0}} \cdot
    \tensor*[^{L}]{\mathbf{P}}{_{i}}
    $}
\put(44,38){\small $\mathbf{n}_{i}$}
\put(37,53){\small $\mathbf{c}_{i}$}
\put(74,28){\small $E_k$}
\end{overpic}
\vspace{-3mm}
\caption{After projecting the point $\tensor*[^{L}]{\mathbf{P}}{_{i}}$ on the 3D edge to the event camera imaging plane at time $t_k$, we select the nearest $m$ event pixels (blue dots) on the $I_k$ image and calculate the center ${\mathbf{c}}{_{i}}$ and normal vector $\mathbf{n}_{i}$.}
\label{fig:3d_point_to_event_image_projection}
\end{figure}
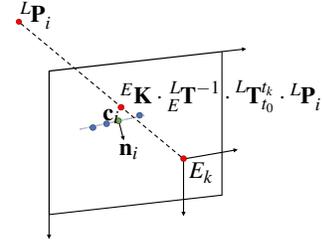

\subsection{Point Cloud to Events Optimization}
\label{sec:optimize}
In our approach, we aim to associate edges extracted from the static point cloud with events. To accomplish this, we warp the events $\mathcal{E}_k$ into the event image $I_k$ with the parameter $\mathbf{\omega}^{*}_k$ estimated in~\prettyref{sec:contrast_maximization}. As illustrated in \prettyref{fig:3d_point_to_event_image_projection}, each point $P_i$ in the edge point cloud $\mathcal{P}_e$ can be projected onto the event camera's imaging plane to obtain the corresponding projection point
\begin{equation}
    \begin{aligned}
    \tensor*[^{E}]{\mathbf{p}}{_{i}}=
    \tensor*[^{E}]{\mathbf{K}}{} \cdot
    \tensor*[^{L}_{E}]{\mathbf{T}}{^{-1}} \cdot
    \tensor*[^{L}]{\mathbf{T}}{^{t_k}_{t_0}} \cdot
    \tensor*[^{L}]{\mathbf{P}}{_{i}}.
    \end{aligned}
    \label{eq:projection}
\end{equation}
This projection process involves several parameters, including the intrinsic parameters of the event camera $\tensor*[^{E}]{\mathbf{K}}{}$, the current extrinsic parameters $\tensor*[^{L}_{E}]{\mathbf{T}}{}$, the transformation of LiDAR from $t_0$ to $t_k$ denoted by $\tensor*[^{L}]{\mathbf{T}}{^{t_k}_{t_0}}$, and the homogeneous coordinates of the point $\tensor*[^{L}]{\mathbf{P}}{_{i}}$.


To ensure that the projection point $\tensor*[^{E}]{\mathbf{p}}{_{i}}$ is accurately placed on the edge formed by the neighboring event pixels, we first search for the $m$ closest event pixels of this projection in $I_k$. From these pixels, we calculate their mass center $\tensor*[]{\mathbf{c}}{_{i}}$ and normal vector $\mathbf{n}{_i}$, respectively. We can then compute the distance of $\tensor*[^{E}]{\mathbf{p}}{_{i}}$ to the edge as the inner product of the normal vector and the difference between the projection point and the mass center, as expressed in \prettyref{eq:point_to_line_dist}.
\begin{equation}
    \begin{aligned}
    d=\langle \tensor*[]{\mathbf{n}}{_{i}},
    \tensor*[^{E}]{\mathbf{p}}{_{i}}-\tensor*[]{\mathbf{c}}{_{i}}\rangle.
    \end{aligned}
    \label{eq:point_to_line_dist}
\end{equation}

We can sample $N$ points on the edge of the point cloud at each sampled time interval, and they should satisfy the above point-to-edge constraint. 
For $K$ sampled time interval, or $K$ poses, we can define the cost function as
\begin{equation}
    \begin{aligned}
    f(\tensor*[^{E}_{L}]{\mathbf{T}}{})=\sum_{k=1}^{K}\sum_{i=1}^{N}
    \langle \tensor*[]{\mathbf{n}}{_{i}},
    \tensor*[^{E}]{\mathbf{K}}{} \cdot
    \tensor*[^{L}_{E}]{\mathbf{T}}{^{-1}} \cdot
    \tensor*[^{L}]{\mathbf{T}}{^{t_k}_{t_0}} \cdot
    \tensor*[^{L}]{\mathbf{P}}{_{i}}
    -\tensor*[]{\mathbf{c}}{_{i}}\rangle.
    \end{aligned}
    \label{eq:cost_function}
\end{equation}


We employed a gradient-based non-linear optimization algorithm to minimize the error described in~\prettyref{eq:cost_function} and determine the extrinsic transform $\tensor*[^{L}_{E}]{\mathbf{T}}{}$. We iteratively solved for $\tensor*[^{L}_{E}]{\mathbf{T}}{}$ until convergence, which yielded the calibrated extrinsic parameters. To implement this approach, we chose to use the Ceres Solver\footnote{http://ceres-solver.org/index.html}.

\section{Experiments}
\subsection{Experimental Setup}

Our experimental sensor setup comprises an iniVation DAVIS346 event camera and a Livox Avia solid-state LiDAR, as illustrated in  Fig. \ref{fig:paper_purpose}. In our method for extrinsic calibration, we only use the event sensor in DAVIS346. The frame-based sensor in DAVIS346 is used for providing reference results as will be described later.
In all experiments, we report the deviation of the estimated extrinsic transform, $^{L}_{E}\mathbf{T}$, from the nominal extrinsic transform which is $(0,0,0)$ for translation and $(0, \pi/2, \pi/2)$ for rotation (ZYX Euler Angle).

Prior to data collection, we calibrated the event camera's intrinsic parameters and lens distortion parameters. We conducted data collection in six scenes, where both the event camera and LiDAR were stationary for $\sim$20 seconds before engaging in motion for $\sim$15 seconds. The movement was a handheld 3D rotational motion centered on the event camera. Throughout the six scenes, the event camera and LiDAR maintained consistent extrinsic parameters.

\begin{figure}[t]
        \vspace{2.5mm}
	\centering
	{
		\begin{minipage}{\linewidth} 
            \includegraphics[width=\textwidth]{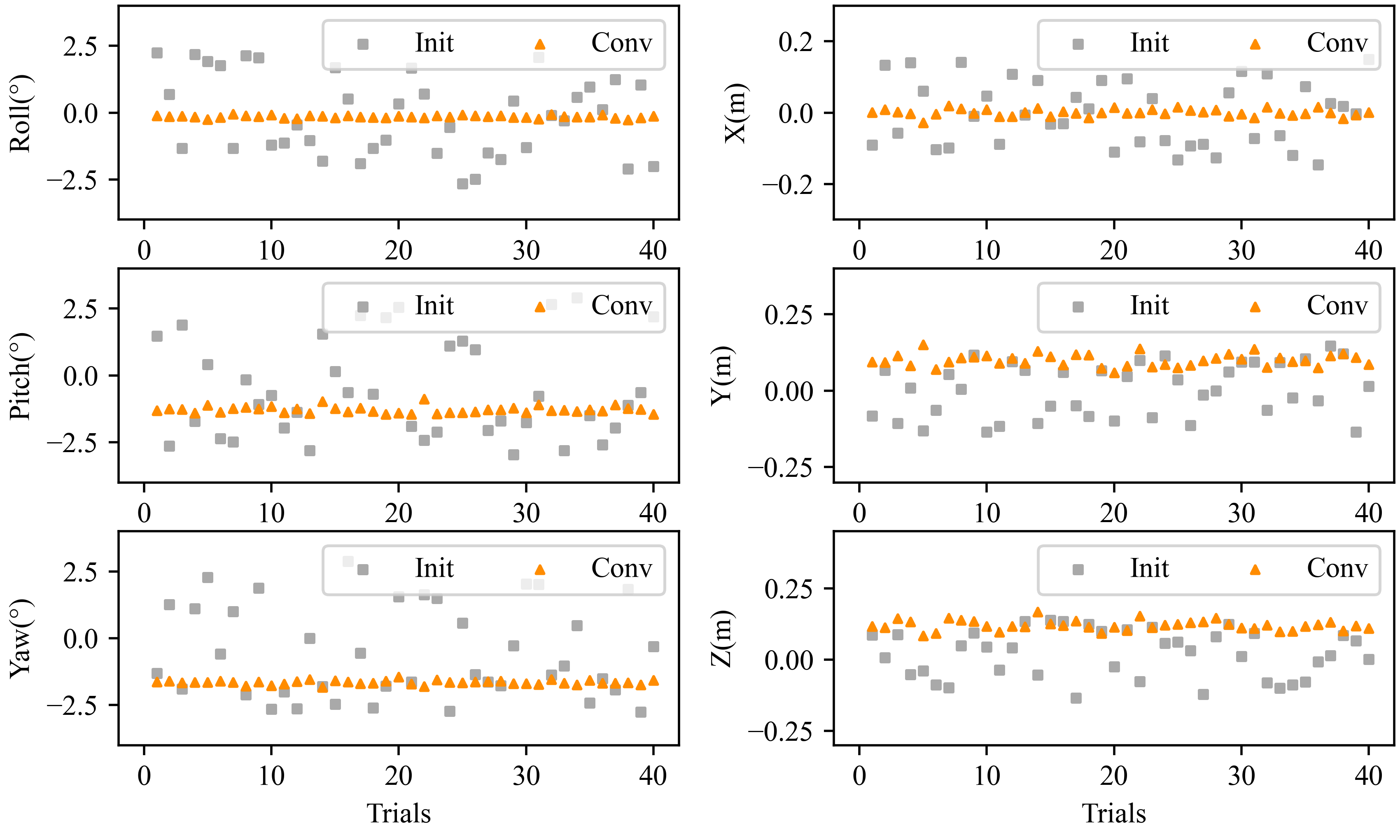} \\
              \vspace{-4mm}
		\end{minipage}}
	\caption{Random initializations (in grey) and converged solutions (in orange) for extrinsic parameters on scene 4 are presented. It is noteworthy that the demonstrated extrinsic parameters are devoid of their nominal parts.} 
 \label{fig:robustness}
\end{figure}

\subsection{Robustness Analysis}
\label{sec:robustness}

To assess the robustness of our method against different initializations, we evaluated the extrinsic calibration process with various initial values randomly sampled around the nominal extrinsic parameters. In particular, we introduced a $3\degree$ rotation about a random axis to the nominal rotation matrix and added a random translation vector with a length of 0.2m to the nominal translation vector. We conducted 40 trials of the experiment in scene 1, as shown in \prettyref{fig:robustness}, with the gray dots representing the initial values and the colored dots representing the final convergence values. Our results demonstrate that our method is highly resilient to different initial values, with the final calibration results remaining consistent across multiple trials.

\begin{figure}[t]
\vspace{2mm}

	\centering
	{
		\begin{minipage}{\linewidth} 
            \includegraphics[width=\textwidth]{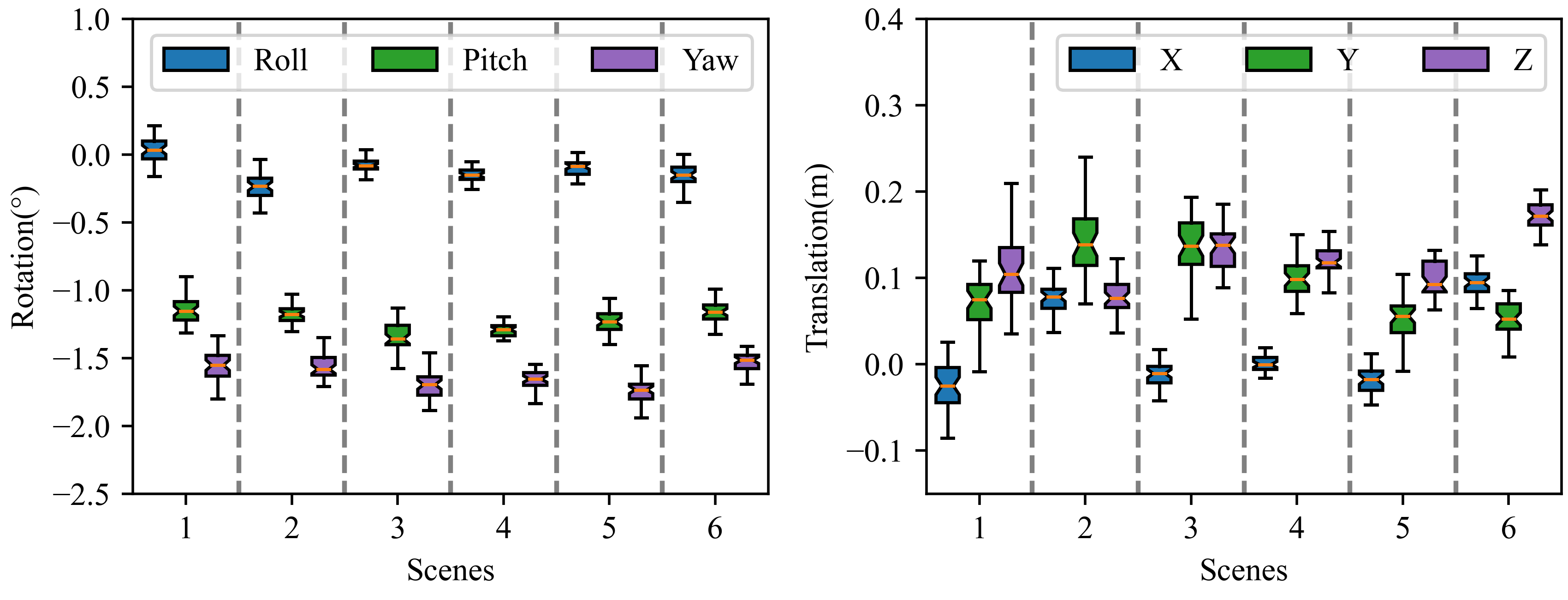} \\
            \vspace{-4mm}
		\end{minipage}}
	\caption{The box plots encapsulate the statistical details, comprising maximum, third quartile, median, first quartile, and minimum, of extrinsic parameters over six distinct scenes. It is noteworthy that the demonstrated extrinsic parameters are devoid of their nominal parts.} 
    \label{fig:consistency}
\end{figure}

We also conducted the above calibration experiment (with 40 trials) on all six sets of experimental data using random initial values according to the randomization process described above to evaluate the robustness in different scenes.
The statistical details of the convergence values of the algorithm across the different scenarios are presented in \prettyref{fig:consistency}. Our method produced consistent calibration results across the different scenarios. We observed that the richness of edges within different scenes influenced the calibration results. In particular, scenes with clearer edges led to better outcomes. This can be attributed to the fact that our method relies on edge correspondences, and clearer edges provide more accurate information for the calibration process. 


\begin{figure}[t]
        \vspace{0.33cm}
	\centering
	{
		\begin{minipage}{\linewidth} 
            \includegraphics[width=\textwidth]{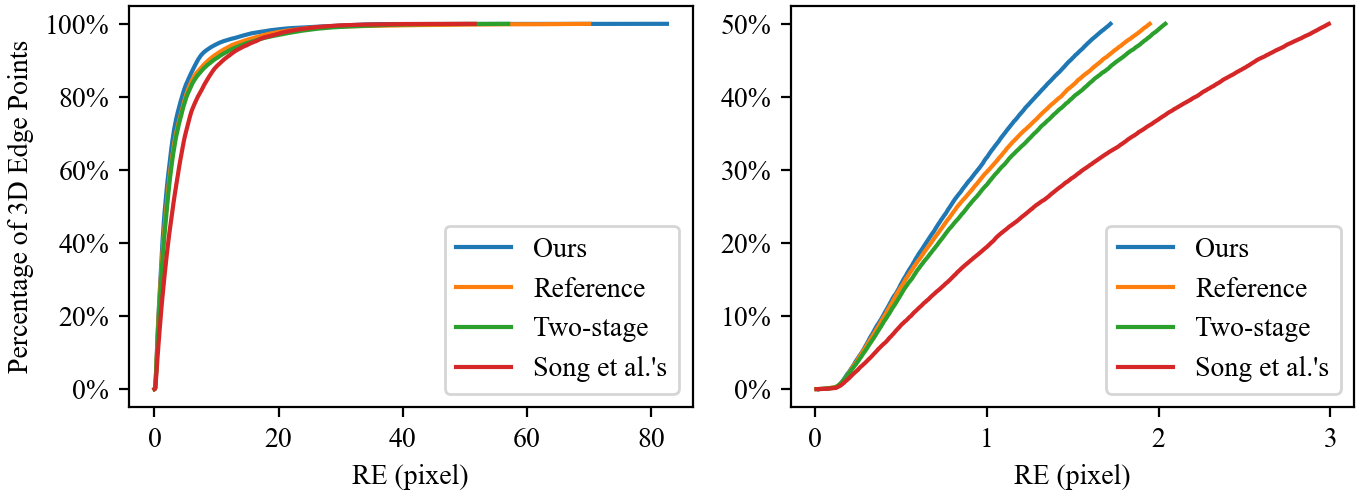} \\
            \vspace{-4mm}
		\end{minipage}}
	\caption{All 3D edge points' RE and the top $50\%$ points' RE under the extrinsic parameters.} 
    \label{fig:dist_distribution}
\end{figure}
\subsection{Performance Comparison}
We conducted a comparative analysis of our method with two existing methods, namely Song \textit{et al.}'s method ~\cite{song2018calibration} and the two-stage method. 
Song \textit{et al.}'s method involves the use of a calibration board with circular holes and a continuously blinking display. The method detects circles from the accumulated events image and the point cloud and then calibrates the translation and rotation parameters in stages. In the two-stage approach, an additional frame-based camera is introduced, and the event-to-video reconstruction~\cite{rebecq2019high, Muglikar2021CVPR} is employed to obtain the frame image of the event camera. The extrinsic parameters of the event camera and introduced frame-based camera are calibrated using the standard stereo camera calibration method. Subsequently, the frame-based camera and LiDAR extrinsic parameter calibration method~\cite{zhou2018automatic} is used to calibrate the extrinsic parameters between the introduced frame-based camera and LiDAR, and finally, the two extrinsic parameters are combined to obtain the extrinsic parameters between the event camera and LiDAR. 
Finally, as a reference, we leverage the frame-based camera in the DAVIS346 event camera for calibrating the event sensor and the LiDAR. 
As the optical path of the DAVIS346 event camera is shared by both its frame and event sensors, the extrinsic parameters between the frame-based sensor and the LiDAR are equivalent to that of the event sensor and the LiDAR. We use this result as a reference for evaluation. To calibrate the extrinsic parameters between the frame-based sensor and the LiDAR, we utilized the standard calibration board under 20 different poses. 

\begin{figure}[t]
    \subfloat[Accumulated events image with varying luminance]{%
      \begin{overpic}[width=0.475\linewidth]{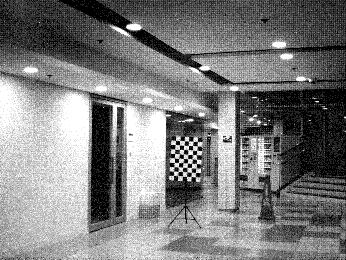}
      \end{overpic}}
    \hfill    
    \subfloat[Point cloud reflectivity image]{%
      \begin{overpic}[width=0.475\linewidth]{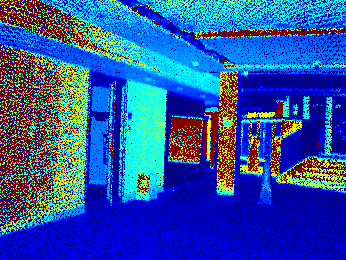}
      \end{overpic}}
    \\
    \subfloat[Our method]{
      \begin{overpic}[width=0.46\linewidth]{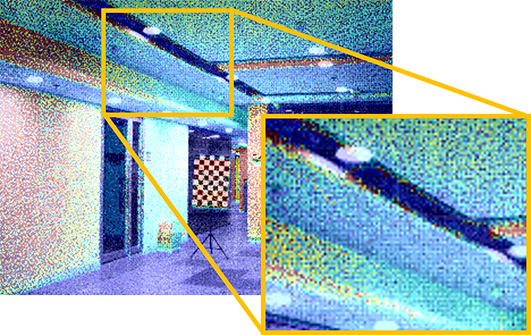}
      \end{overpic}}
    \hspace{1.9mm}
    \subfloat[Reference method]{
      \begin{overpic}[width=0.46\linewidth]{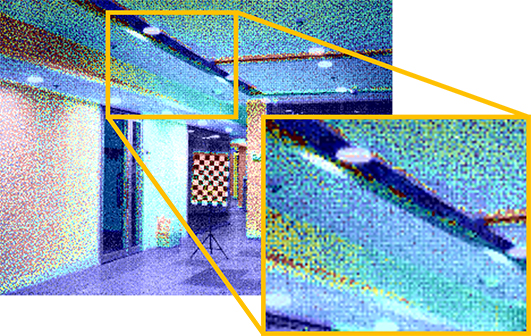}
      \end{overpic}}
    \\
    \subfloat[Two stage method]{
      \begin{overpic}[width=0.46\linewidth]{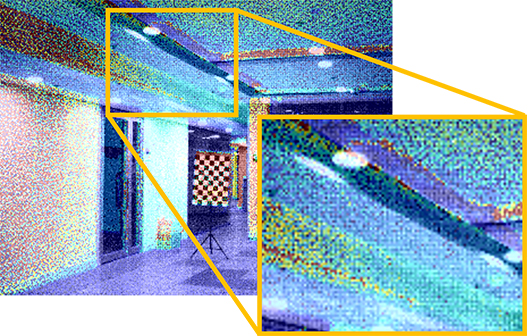}
      \end{overpic}
      }
    \hspace{1mm}
    \subfloat[Song \textit{et al.}'s method]{
      \begin{overpic}[width=0.46\linewidth]{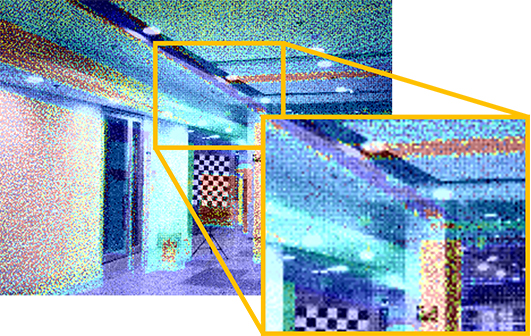}
      \end{overpic}
    }
    \\
    \caption{Visual comparison among reprojection results by our method (c), the reference method using DAVIS346 frame-based camera (d), the two-stage method (e) and Song \textit{et al.}'s method (f). Our method performs \textit{on par} with the reference method.}
    \label{fig:visual_comparison}

\end{figure}

\begin{figure}[t]
    \subfloat[Static environment point cloud and segmented moving object point cloud (in white)]{%
      \begin{overpic}[width=0.475\linewidth]{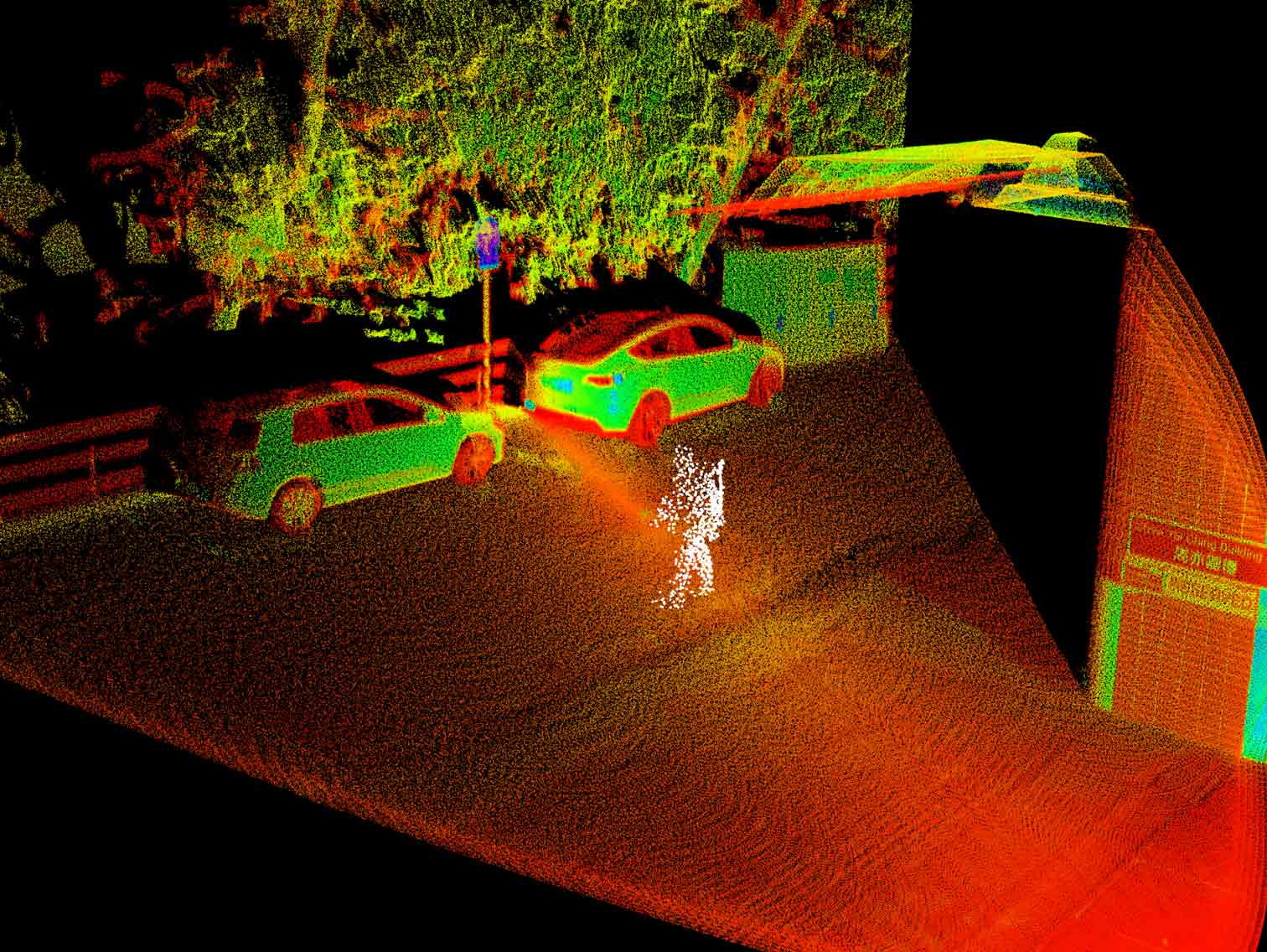}
      \end{overpic}}
    \hfill    
    \subfloat[Event image and overlayed point cloud projection]{%
      \begin{overpic}[width=0.475\linewidth]{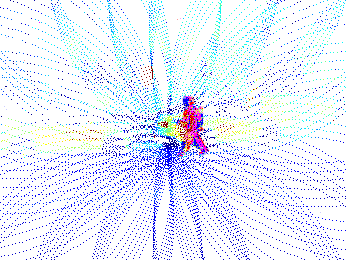}
      \end{overpic}}
    \\
    \caption{The event camera and LiDAR are fixed within the environment. Leveraging calibrated extrinsic parameters, we can segment the event-related point cloud from the overall point cloud.}
    \label{fig:application}
\end{figure}

\renewcommand\arraystretch{1.5}
\begin{table}[ht]
\centering
\caption{PPRE comparison among our calibration, reference, the two-stage method and Song \textit{et al.}'s method}
\begin{tabular}[t]{lcccc}
\toprule
&Ours            &Reference      &Two-stage   &Song \textit{et al.}'s   \\
\midrule
PPRE    &\textbf{3.161}         &3.726          &3.922              &4.691      \\
\bottomrule
\end{tabular}
\label{tab:numeric_comparison}
\end{table}

To evaluate the effectiveness of our calibration method, we conducted experiments in a static environment with varying luminance. We directly accumulated events to obtain event image and utilized the extrinsic parameters obtained by different methods to project the point cloud onto the image plane of the event camera to obtain the reflectivity image of the projected point cloud. Then, we overlaid the reflectivity image with the event image, as shown in \prettyref{fig:visual_comparison}, to visually check the reprojection quality. 
Our method exhibited the best alignment among the evaluated methods, and the calibration quality closely approximated the reference approach that uses the frame-based sensor of DAVIS346 for calibration. 


We collected a total of 60 poses, uniformly distributed among 6 scenes, and utilized different extrinsic parameters to project the point cloud edges onto the imaging plane of the event camera. Using \prettyref{eq:point_to_line_dist}, we determined the reprojection error (RE) by calculating the distance between the edge points' reprojections and the adjacent event pixels. While appropriate extrinsic parameters ensure that most edge points are situated near adjacent events, imperfect correspondences between the point cloud edges and events may result in large RE values for certain points, even with ideal extrinsic parameters. \prettyref{fig:dist_distribution} presents the distribution of RE values for all points and the top $50\%$ of points under different extrinsic parameters. Our method's extrinsic parameters led to a significant reduction in RE values for most points. Additionally, we computed the per-point reprojection error (PPRE) for all points, as demonstrated in \prettyref{tab:numeric_comparison}. The results reveal that our method produces the smallest reprojection error compared to other methods.

\subsection{Application}
In \prettyref{fig:application}, we demonstrate the practical utility and efficacy of our event camera and LiDAR extrinsic parameter calibration. With the devices fixed in the environment and calibrated extrinsic parameters applied, we can effectively segregate the event-related point cloud from the overall point cloud.

\subsection{Limitations}

Our method may exhibit reduced calibration accuracy in situations with sparse or indistinct edges, resulting in less precise extrinsic parameters and higher reprojection errors. Moreover, when the sensors are significantly separated~\cite{berrio2021camera}, establishing correspondence between event and LiDAR data becomes challenging, impacting calibration precision.
\section{Conclusions}

Our paper proposes an innovative method that utilizes edge correspondences to achieve the extrinsic calibration of a dyad of a LiDAR and an event camera. This approach eliminates the requirement for specialized calibration board or equipment to change the luminance, allowing us to complete the calibration process in a general environment. As event camera technology evolves, our method has the potential to be applied in a variety of applications involving event cameras and LiDAR fusion.




\bibliographystyle{IEEEtran}
\bibliography{IEEEabrv,references}

\end{document}